\documentclass[letterpaper, 10 pt, conference]{ieeeconf}  
\usepackage{caption}
\usepackage{subcaption}
\usepackage{graphicx}
\usepackage{hyperref}
\usepackage{amsmath}
\usepackage{amssymb}
\usepackage[table]{xcolor}
\usepackage{multirow}
\usepackage[utf8]{inputenc}

\IEEEoverridecommandlockouts                              

\title{\LARGE \bf
TransUPR: A Transformer-based Plug-and-Play Uncertain Point Refiner For LiDAR Point Cloud Semantic Segmentation
}

\author{Zifan Yu$^{1}$, Meida Chen$^{2}$, Zhikang Zhang$^{1}$, Suya You$^{3}$, Raghuveer Rao $^{3}$, Sanjeev Agarwal$^{4}$ and Fengbo Ren$^{1}$
\thanks{$^{1}$The authors are with the Parallel Systems and Computing Laboratory, Arizona State University, Tempe, Arizona, USA
        {\tt\small \{zifanyu, zhikang.zhang, fren5\}@asu.edu}}%
\thanks{$^{2}$Meida Chen is with the Institute for Creatives Technologies, University of Southern University, Los Angeles, California, USA
        {\tt\small mechen@ict.usc.edu}}%
\thanks{$^{3}$Suya You and Raghuveer Rao are with U.S. DEVCOM Army Research Laboratory, Adelphi, Maryland, USA
{\tt \small \{suya.you, raghuveer.m.rao\}.civ@army.mil}}
\thanks{$^{3}$Sanjeev Agarwal is with U.S. DEVCOM Army C5ISR Center, Fort Belvoir, Virginia, USA
{\tt \small sanjeev.agarwal.civ@army.mil}}
}

\begin{document}

\maketitle
\thispagestyle{empty}
\pagestyle{empty}

\begin{abstract}

Common image-based LiDAR point cloud semantic segmentation (LiDAR PCSS) approaches have bottlenecks resulting from the boundary-blurring problem of convolution neural networks (CNNs) and quantitation loss of spherical projection. In this work, we propose a transformer-based plug-and-play uncertain point refiner,  i.e., TransUPR, to refine selected uncertain points in a learnable manner, which leads to an improved segmentation performance. Uncertain points are sampled from coarse semantic segmentation results of 2D image segmentation where uncertain points are located close to the object boundaries in the 2D range image representation and 3D spherical projection background points. Following that, the geometry and coarse semantic features of uncertain points are aggregated by neighbor points in 3D space without adding expensive computation and memory footprint. Finally, the transformer-based refiner, which contains four stacked self-attention layers, along with an MLP module, is utilized for uncertain point classification on the concatenated features of self-attention layers. As the proposed refiner is independent of 2D CNNs, our TransUPR can be easily integrated into any existing image-based LiDAR PCSS approaches, e.g., CENet. Our TransUPR with the CENet achieves state-of-the-art performance, i.e., 68.2\% mean Intersection over Union (mIoU) on the Semantic KITTI benchmark, which provides a performance improvement of 0.6\% on the mIoU compared to the original CENet. 

\end{abstract}

\section{INTRODUCTION}

\begin{figure}[t]
\begin{center}
\includegraphics[width=\linewidth]{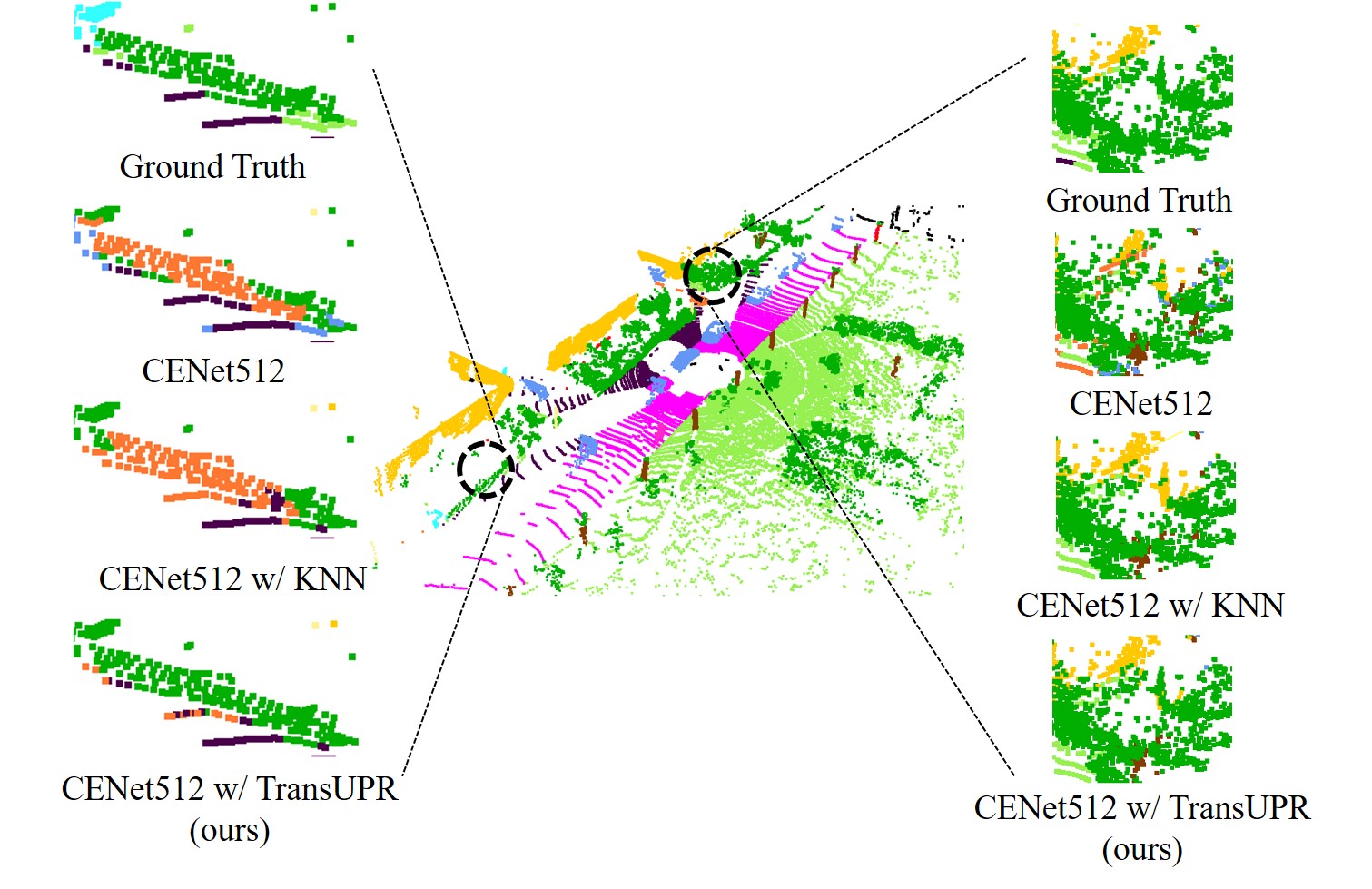}
\end{center}
   \caption{Visualization results of the common KNN refiner and our proposed TransUPR. The KNN refiner may rectify some points that are correctly predicted by CNN backbones, e.g., the vegetation on the left column is incorrectly segmented as "other-structure" by the CENet512 but the KNN refiner cannot correct such errors and even enlarge the area with error segmented points. }
\label{fig:p1}
\end{figure}

Accurate LiDAR point cloud semantic segmentation (LiDAR PCSS) is a fundamental task for robotics and self-driving car and remains challenging when LiDAR point clouds are disordered and sparse. Image-based LiDAR PCSS approaches \cite{squeezeseg, rangenet++, squeezesegv2, squeezesegv3, salsanext} are strong competitors to point-based \cite{pointnet++, randla, kpconv} and partition-based \cite{cylinder3d, point2voxel, minkowski,STPLS3D} LiDAR PCSS methods in real-time applications. However, as a result of the boundary-blurring problem of CNNs and the quantitation loss of spherical projection, image-based PCSS approaches often produce inaccurate labels for points located on the boundary of objects and areas that are far from the foreground points. \cite{salsanext, fidnet} alleviates the boundary-blurring problem and improves segmentation accuracy by applying domain-specific network architecture modifications and loss functions to CNNs borrowed from other 2D vision tasks. \cite{cenet, lite} achieves state-of-the-art performance by utilizing a boundary loss to handle blurred objects caused by upsampling and downsampling. The GPU-accelerated K-Nearest Neighbors (KNN) refiner \cite{rangenet++, salsanext, fidnet, cenet} is a common post-processing technique of the above-mentioned image-based methods to deal with the quantitation loss of spherical projection. The background points, i.e. points that are not presented in the projected 2D image representation, are reclassified based on the predicted labels of their neighbor points after the backward spherical projection. However, image-based methods with a common and non-learnable KNN refiner still fall behind state-of-the-art partition-based \cite{point2voxel} and hybrid methods \cite{rpvnet, fusionnet, 3dmininet, af2-s3net} in terms of segmentation accuracy, and the widely-used KNN refiner may incorrectly rectify points (see Fig. \ref{fig:p1}). 

\begin{figure*}[h]
\begin{center}
\includegraphics[width=\linewidth]{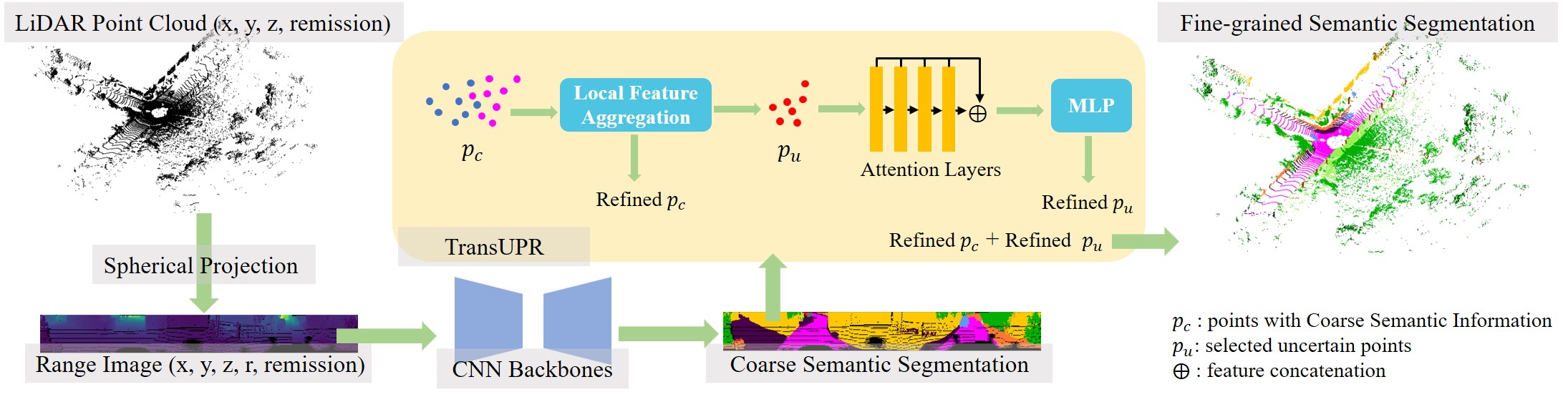}
\end{center}
   \caption{The Pipeline of TransUPR.}
\label{fig:p2}
\end{figure*}

In this paper, we propose a transformer-based plug-and-play refiner for image-based LiDAR PCSS to reclassify uncertain points that are misclassified. As shown in Fig \ref{fig:p2}, given coarse semantic segmentation features generated by the CNN backbones, e.g., CENet\cite{cenet}, the uncertain points are selected in two ways: (i)the points with a lower probability of coarse semantic segmentation labels since these points are highly likely located near the object boundaries; (ii) the points in the background, i.e., points blocked by foreground points during spherical projection and are far away from foreground points. We then use the geometry (e.g., coordinates) along with the locally augmented semantic segmentation features to train our designed learnable transformer-based refiner, which contains four concatenated self-attention layers and an MLP module for uncertain point classification. We benchmark our proposed refiner on Semantic KITTI \cite{semantickitti} with three typical image-based LiDAR PCSS methods and quantitatively and qualitatively compare the performance of common KNN refiner and our TransUPR. 

The contributions of this paper are as follows: 1) We propose a generic transformer-based uncertain point refiner for image-based LiDAR PCSS, which is plug-and-play for all existing image-based CNN backbones to achieve improved performances 2) By integrating with the proposed refiner, the previous state-of-the-art approach\cite{cenet} achieves even higher performance, i.e., from 67.6\% to 68.2\% mIoU, on Semantic KITTI\cite{semantickitti}, which is strong empirical evidence for potentials of our approach in improving the performance of image-based LiDAR PCSS. 3) We draw an insight that focusing on uncertain points with locally aggregated geometry and coarse semantic feature is an efficient way to combine image-based LiDAR PCSS methods with point-based approaches, which break the computation and memory footprint burdens for applying transformer to large-scale point clouds.

\section{Related Work}

Solving LiDAR PCSS with image representation is able to leverage the well-defined and state-of-the-art CNN architectures borrowed from other 2D vision tasks as feature extraction backbones \cite{rangenet++, salsanext, cenet, calib}. Wu et al. \cite{squeezeseg} first proposed the idea to represent sparse and disordered frontal LiDAR point clouds in a dense and grid-based 2D representation, i.e., range image, which largely reduces the redundancy of the 3D voxel-based representation and the computation needed for 3D CNNs. Since the only difference between the range image and the natural image is the number of channels, well-defined CNN architectures such as SqueezeNet \cite{squeezenet} are able to be easily applied to extract features from range images with minor domain-specific modifications. The proposed SqueezeSeg \cite{squeezeseg} serves along with a conditional random field (CRF) which is widely used to smooth the segmentation results for natural image-related segmentation. However, the benefits of the CRF are limited for LiDAR PCSS since a large part of the background points is still affected by the blurring-boundary problem, and the problem is even magnified by the quantitation loss of spherical projection. Milioto et al. \cite{rangenet++} utilized a GPU-enabled KNN refiner to directly post-process the segmented point clouds in the 3D space, which remains a common post-processing technique for image-based LiDAR PCSS approaches; The authors also extended the range image representation to the full LiDAR scan instead of the frontal $90$ degrees of the scan and apply a deeper CNN with residual connections for feature extraction to increase the model capacity \cite{rangenet++, resnet}. Later works such as \cite{salsanext, fidnet, cenet, lite} achieve higher segmentation performance by applying domain-specific network modifications, e.g., dilated convolution layer and pixel-shuffle layer, and boundary loss functions. However, the GPU-enabled non-learnable KNN refiner is still the only widely utilized post-processing technique. Although \cite{fidnet} proposes the nearest label assignment (NLA) refiner, it can be viewed as a special case of KNN refiner when a limited search space for neighbor points is given. The improvement brought by NLA refiner on the Semantic KITTI validation set is also limited, i.e., $0.2\%$ improvement on mIoU. In this paper, we validate the performance and correctness of our proposed TransUPR by leveraging the existing image-based LiDAR PCSS methods \cite{salsanext, fidnet, cenet} as the backbones to generate coarse semantic segmentation and apply post-processing techniques separately, i.e., KNN refiner and our TransUPR.

Moreover, extracting features from multiple representations to boost the performance of LiDAR PCSS attracts attention thanks to the recently proposed 3D sparse convolution technique \cite{minkowski}, which alleviates the harsh computing requirement needed for 3D CNNs. Zhang et al., \cite{fusionnet} proposed a voxel-based MLP with a structure named mini-PointNet, which is able to learn point-wise features from points among all the neighbor voxels with low search effort, and the point-wise features can avoid the quantitation loss of voxelization. Alonso, et al., \cite{3dmininet} reduces the expensive computation needed among common point-wise methods with the help of regular spherical projection for neighbor point searching, and a point-based learnable network was utilized to learn a 2D representation from grouped 3D points. The visual features extracted from the learned 2D representation are combined with the range image for image-based LiDAR PCSS. Cheng, et al., \cite{af2-s3net} proposed an attentive fusion module to merge the extracted point-based, small-scale-based, and large-scale-based features. Then, the key features are highlighted by the following adaptive feature selection module. However, all the mentioned hybrid methods focus on combining multiple representations at the feature extraction and formation level. In this paper, we rethink this problem and provide a feasible way to mix image-based methods with point-based methods in a sequential manner. Each module in our pipeline is plug-and-play and can be replaced by any other alternatives. 

\begin{table*}[h]
 \centering
 \scalebox{0.8}{
 \begin{tabular}{c|ccccccccccccccccccc|c}
 \hline
Methods & \textit{\rotatebox[origin=c]{90}{car}} & \textit{\rotatebox[origin=c]{90}{bicycle}} & \textit{\rotatebox[origin=c]{90}{motorcycle}}& \textit{\rotatebox[origin=c]{90}{truck}} & \textit{\rotatebox[origin=c]{90}{other-vehicle}} & \textit{\rotatebox[origin=c]{90}{person}} & \textit{\rotatebox[origin=c]{90}{bicyclist}} & \textit{\rotatebox[origin=c]{90}{motorcyclist}} &  \textit{\rotatebox[origin=c]{90}{road}} &  \textit{\rotatebox[origin=c]{90}{parking}} &  \textit{\rotatebox[origin=c]{90}{sidewalk}} &  \textit{\rotatebox[origin=c]{90}{other-ground}} &  \textit{\rotatebox[origin=c]{90}{building}} & \textit{\rotatebox[origin=c]{90}{fence}} &  \textit{\rotatebox[origin=c]{90}{vegetation}} & 
\textit{\rotatebox[origin=c]{90}{trunk}} &  \textit{\rotatebox[origin=c]{90}{terrain}} &
\textit{\rotatebox[origin=c]{90}{pole}} & \textit{\rotatebox[origin=c]{90}{traffic-sign}} & \textit{\textbf{\rotatebox[origin=c]{90}{mIoU}}}\\
 \hline \hline
SqueezeSeg\cite{squeezeseg} &  68.8 & 16.0 & 4.1 & 3.3 & 3.6 & 12.9 & 13.1 & 0.9 & 85.4 & 26.9 & 54.3 & 4.5 & 57.4 & 29.0 & 60.0 & 24.3 & 53.7 & 17.5 & 24.5  & 29.5\\
SqueezeSeg-CRF\cite{squeezeseg} & 68.3 & 18.1 & 5.1 & 4.1 & 4.8 & 16.5 & 17.3 & 1.2 & 84.9 & 28.4 & 54.7 & 4.6 & 61.5 & 29.2 & 59.6 & 25.5 & 54.7 & 11.2 & 36.3  & 30.8 \\
SequeezeSegV2 \cite{squeezesegv2} & 81.8 & 18.5 & 17.9 & 13.4 & 14.0 & 20.1 & 25.1 & 3.9 & 88.6 & 45.8 & 67.6 & 17.7 & 73.7 & 41.1 & 71.8 & 35.8 & 60.2 & 20.2 & 36.3& 39.7\\
SequeezeSegV2-CRF \cite{squeezesegv2} & 82.7 & 21.0 & 22.6 & 14.5 & 15.9 & 20.2 & 24.3 & 2.9 & 88.5 & 42.4 & 65.5 & 18.7 & 73.8 & 41.0 & 68.5 & 36.9 & 58.9 & 12.9 & 41.0 & 39.6\\
SequeezeSegV3 \cite{squeezesegv3} & 92.5 & 38.7 & 36.5 & 29.6 & 33.0 & 45.6 & 46.2 & 20.1 & 91.7 & 63.4 & 74.8 & 26.4 & 89.0 & 59.4 & 82.0 & 58.7 & 65.4 & 49.6 & 58.9 & 55.9\\
RangeNet++ \cite{rangenet++} & 91.4 & 25.7& 34.4 &25.7& 23.0& 38.3 &38.8 &4.8 &91.8& 65.0& 75.2& 27.8& 87.4& 58.6& 80.5& 55.1& 64.6& 47.9& 55.9& 52.2\\
\hline
SalsaNext \cite{salsanext} & 91.9 & 46.0& 36.5& 31.6& 31.9& 60.9& 61.7& 19.4& 91.9 & 63.6 & 76.1 & 28.9 & 89.8 & 63.1 & 82.1 & 63.5 & 66.4 & 54.0 & 61.4  & 59.0 \\
SalsaNext \cite{salsanext} + \textbf{TransUPR} & \cellcolor{green}{92.4} & 45.3 & \cellcolor{green}{37.0} & 30.9& 
31.5& \cellcolor{green}{61.9} & \cellcolor{green}{63.2} & 16.2 & \cellcolor{green}{\textbf{92.0}} & \cellcolor{green}{64.0} & \cellcolor{green}\textbf{76.2} & 28.9 & \cellcolor{green}\textbf{90.3} & \cellcolor{green}\textbf{63.9} & \cellcolor{green}{83.0} & \cellcolor{green}{66.0} & \cellcolor{green}{67.2} & \cellcolor{green}{55.2} & 60.3  & \cellcolor{green}{59.2} \\
FIDNet\cite{fidnet} & 92.3 & 45.6 & 41.9 & 28.0 & 32.6 & 62.1 & 56.8 & 30.6 & 90.9 & 58.4 & 74.9 & 20.5 & 88.6 & 59.1 & \textbf{83.1} & 64.6 & 67.8 & 53.1 & 60.1 & 58.5\\
FIDNet\cite{fidnet} + \textbf{TransUPR} &  \cellcolor{green}{92.6} & 43.6 & \cellcolor{green}{42.0} & \cellcolor{green}{29.7} & \cellcolor{green}{32.7} & \cellcolor{green}{63.3} & \cellcolor{green}{58.8} & 27.0 & 90.8 & \cellcolor{green}{58.5} & 74.5 & 20.2 & \cellcolor{green}{88.7} & \cellcolor{green}{59.9} & 82.9 & \cellcolor{green}\textbf{66.5} & 67.7 & \cellcolor{green}{54.1} & \cellcolor{green}{61.0}& \cellcolor{green}{58.7} \\
CENet512\cite{cenet} & 93.0 & 51.2 & 67.2 & 58.4 & 62.4 & 69.1 & 70.0 & \textbf{54.3} & 90.3 & 70.0 & 74.5 & 40.8 & 88.6 & 61.3 & 81.6 & 63.5 & 69.2 & 55.5 & \textbf{63.1}  & 67.6 \\
CENet512\cite{cenet} + \textbf{TransUPR} & \cellcolor{green}\textbf{93.3} & \cellcolor{green}\textbf{53.6} & \cellcolor{green}\textbf{68.1} & \cellcolor{green}\textbf{58.6} & \cellcolor{green}\textbf{63.7} & \cellcolor{green}\textbf{70.0} & \cellcolor{green}\textbf{70.6} & 52.3 & \cellcolor{green}{90.4} & \cellcolor{green}\textbf{70.2} &\cellcolor{green}{74.8} & \cellcolor{green}\textbf{42.1} & \cellcolor{green}{89.1} & \cellcolor{green}{62.8} & \cellcolor{green}{82.3} &  \cellcolor{green}{64.6} & \cellcolor{green}\textbf{{69.4}}& \cellcolor{green}\textbf{{56.4}}& 63.0 & \cellcolor{green}\textbf{68.2} \\
\hline
\end{tabular}}
 \caption{Quantitative Results Comparison on Semantic KITTI Test Set (Sequences 11 to 21). If no specified, scores are given in percentage (\%) in all tables. The input size is $64 \times 2048$ except for CENet-512, which takes range images with a size of $64 \times 512$ as inputs. For the baseline methods without TransUPR, the GPU-enabled KNN refiner is the default post-processing method. (Note: \colorbox{green}{Improved class} and \textbf{best class-wise IoU}).}
 \label{tab:1}
 \end{table*}
 
\section{Methodology}
\label{sec:method}
As shown in Fig. \ref{fig:p2}, for a LiDAR point cloud $P=\mathbb{R}^{N \times 4}$, we first convert each point $p_i = (x, y, z, remission)$ of $P$ into a 2D representation $R$, i.e., range image, with a size of $H \times W \times C$ by spherical projection \cite{squeezeseg, rangenet++}, 
\begin{equation}
\binom{u}{v} = \binom{\frac{1}{2}\left[ 1 - \arctan{(y, x)\pi^{-1}}\right] \times W}{\left[ 1 - \frac{\arcsin(zr^{-1}) + f_u}{f} \right ] \times H},
\end{equation}
where $(u, v)$ are the range image coordinates, $W$ and $H$ is the expected width and height of the target range image, $f = f_u + f_d$ is the vertical field-of-view, and $f_u$ and $f_d$ is the up and down vertical field-of-view, respectively. Each pixel in $R$ contains five channels $(x, y, z, r, remission)$, where $r$, i.e., range, is the Euclidean distance between the point and the sensor origin in the 3D space. One pixel only presents the geometry feature of the foreground projected point with the closest range to the origin. We then utilize existing range image segmentation CNNs as our backbone that takes R as the input to extract coarse semantic segmentation features. Following that, we select the uncertain points and locally aggregate the point features for refinement, and produce point-wise fine-grained semantic labels. We explain the mechanism of our proposed TransUPR module from the following three aspects: (1) the local feature aggregation of geometry features and coarse semantic segmentation features; (2) the strategies of selecting uncertain points for training and testing; (3) the architecture of the learnable transform-based refiner and self-attention layers.  

\subsection{Local Feature Aggregation}
\label{sec:2.1}
We hypothesize that the output of the softmax layer in CNN backbones is the pixel-wise class probabilities, i.e., coarse semantic segmentation features, and \cite{pointrend} points out that pixels with low probabilities are mainly located close to the object boundaries. We back-projected class probabilities of the foreground points to all points projected in the same pixel.  As the points projected into the same 2D pixel have duplicated coarse semantic segmentation features, we aggregate the point semantic features by averaging the class probabilities of its selected $k$ neighbor points, e.g., $k = 7$ for Semantic KITTI \cite{semantickitti}, in the 3D space to make the features distinguishable. Our implementation is based on the original GPU-enabled KNN \cite{rangenet++}, which means the local feature aggregation process does not add harsh computation burdens. Although multiple 3D points may be projected into the identical 2D coordinate, it is possible for them to be far away from each other in the 3D space. Therefore, we concatenate the point geometry attributes $(x, y, z, r, remission)$ and the aggregated coarse semantic features as the uncertain point features for the refiner to increase the capacity of description when the points have similar semantic features. Therefore, a finalized feature vector of a 3D point has a size of $N_{geometry} + N_{classes}$, where $N_{geometry} = 5$ and $N_{classes} = 20$ in Semantic KITTI \cite{semantickitti}. The GPU-enabled local feature aggregation module also generates a coarse point-wise label assignment $refined p_c$ simultaneously. 
 
\subsection{Strategies of Selecting Uncertain Points}
\label{sec:2.2}
Applying the typical transformer \cite{transformer} to the large-scale LiDAR point cloud severely burdens the GPU memory footprint and computation \cite{pointtransformer}. However, we notice that the image-based methods successfully label most of the 3D points, and only a small amount of points are misclassified due to the boundary-blurring problem and quantitation loss of projection. Thus, we localize such points, i.e., uncertain points, to lower the input size for the following transformer-based refiner to avoid memory and computation burdens.  

One strategy is to identify uncertain points in 2D representation, where points are misclassified due to the boundary-blurring problem of CNNs. First, we rank the 2D pixels in ascending order of top-2 class probability differences, where the class probability is the output of the Softmax layer in CNNs. The 3D points projected into the same 2D pixel have identical top-2 class probability differences. Lower top-2 class probability differences imply that the assigned semantic labels are low confidence. Then, we randomly select the top $N_{ru}$, where $N_{ru} = 8192$ for Semantic KITTI \cite{semantickitti}, points with the lowest top-2 class probability differences into the uncertain point pool. 

Another uncertain point localization way is viewing all the background points of spherical projection as the uncertain points in 3D space. However, we observe that most of the background points close to the foreground points belong to the same semantic label as the foreground points in the ground truth. Therefore, we set a cutoff $c_u$ to filter out the background points that are close to the foreground points, and the rest of the background points are the uncertain points. In other words, if the distance, i.e, range,  between a background point and its corresponding foreground points is less than the $c_u$, i.e., $c_u = 1$ for Semantic KITTI\cite{semantickitti}, we put the background point into the uncertain point pool with $N_u$ uncertain points. We randomly select $N_{t}$ points from the uncertain point pool for the refiner training, where $N_{t} = 4096$ for Semantic KITTI. Therefore, given $B$ LiDAR scans, the size of the uncertain point features is $B \times N_{t} \times (N_{geometry} + N_{classes})$ for model training. All points in the uncertain point pool are refined during the model inference so the refiner takes input with a size of $1 \times \lceil \frac{N_{u}}{N_{t}} \rceil \times (N_{geometry} + N_{classes})$ for point-wise fine-grained labeling.

\subsection{Transformer-based Refiner and Attention Layers}

Given an input uncertain point $p_u \in \mathbb{R}^{N_{geometry} + N_{classes}}$, a $d_e$-dimensional feature embedding $F_e \in \mathbb{R}^{d_e}$ is first learned via two fully connected layers, where $d_e = 256$. We modify the typical transformer \cite{transformer}, i.e., na\"ive point cloud transformer \cite{pointtransformer},  with four self-attention layers, where each uncertain point is viewed as a word. As shown in figure \ref{fig:p2},  the inputs of each self-attention layer are the output of the precedent self-attention layer, and the outputs of each self-attention layer are concatenated together for the following uncertain point classification. Let $\mathbf{Q}, \mathbf{K}, \mathbf{V}$ be the \textit{\textbf{query}}, \textit{\textbf{key}}, and \textit{\textbf{value}}. For each attention layer, the $\mathbf{Q}$ and $\mathbf{K}$ are first learned by a shared fully connected layer $W_{p}$, and the $\mathbf{V}$ is learned via another fully connected layer $W_{v}$,
\begin{equation}
    \begin{split}
    (\mathbf{Q}, \mathbf{K}) = W_{p}(F_{in}) \\
    \mathbf{V} = W_{v}(F_{in}),
    \end{split}
\end{equation}
where $F_{in}$ is the input feature embedding. The $\mathbf{Q}$ and $\mathbf{K}$ are utilized to calculate the correlation, i.e., attention weights, by the matrix dot-product, and the attention weights are normalized,

\begin{equation}
   \bar{a}_{i, j} = \frac{\tilde{a}_{i, j}}{\sqrt{d}} = \frac{\mathbf{Q} \cdot \mathbf{K^{T}}}{\sqrt{d}},
\end{equation}
where $d$ is the dimension of the learned feature embedding of the $\mathbf{Q}$ and $\mathbf{K}$. The output feature of a attention layer is the sum of the $\mathbf{V}$ using the attention weights,
\begin{equation}
    F_{out} = \mathbf{A} \cdot \mathbf{V} = softmax(\bar{a}_{i, j}) \cdot \mathbf{V} = \frac{\exp{\bar{a}_{i, j}}}{\sum_{k}\exp{\bar{a}_{i, k}}} \cdot \mathbf{V}.
\end{equation}

The dimension of the output features is $256$ for all four self-attention layers. The concatenated features are fed into the following three fully-connected layers for point classification. The loss function of our TransUPR is the sum of a softmax cross-entropy loss and a \textit{Lov{\'a}sz-Softmax} \cite{lovasz} loss,
\begin{equation}
    L = L_{wce}(y, \hat{y}) + L_{ls}(C), 
\end{equation}
where $y$ is the ground truth label of an uncertain point, and $\hat{y}$ is the predicted label. $C$ is the number of semantic classes for model training. After reclassifying the semantic label of selected uncertain points, we replace the label of uncertain points in refined $p_c$ (see Fig. \ref{fig:p2}) with the label of refined $p_u$.
\begin{figure*}
\begin{center}
\includegraphics[width=0.78\textwidth]{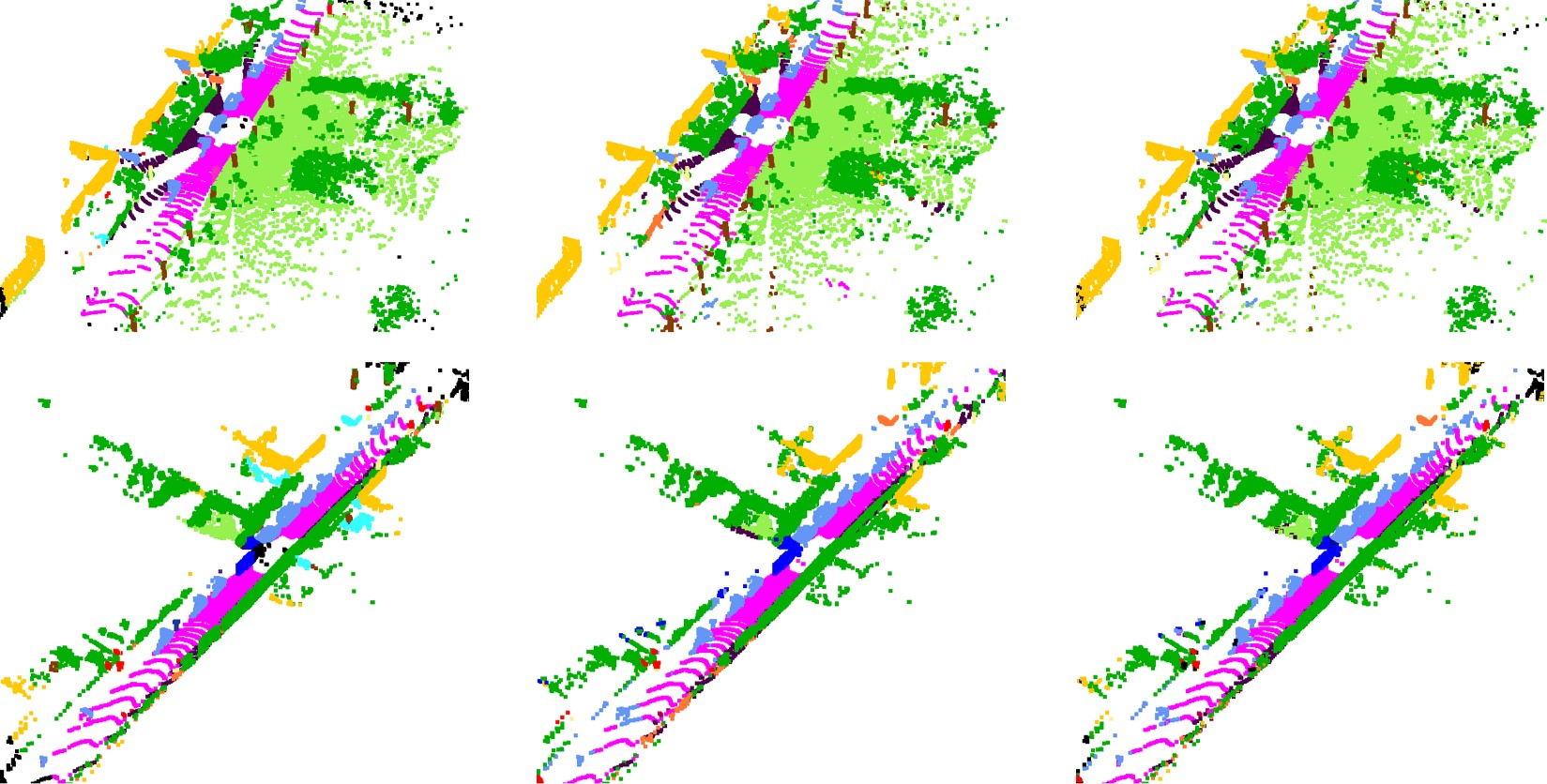}
\end{center}
   \caption{Visualization results on sequence 08. Column 1: point clouds with ground truth labels. Column 2: point clouds with predicted labels of CENet 512 with a KNN refiner. Column 3: point clouds with predicted labels of CENet with our TransUPR.}
\label{fig:p3}
\end{figure*}
\section{Experiments}

\subsection{Dataset and Experiment Settings}

We evaluate the performance of our proposed TransUPR on the large-scale challenging Semantic KITTI dataset\cite{semantickitti} that contains 22 sequences of different scenarios. Over 19k scans (sequences between 00 and 10 except for 08) are used for training, and sequence 08 is used as a validation set. 19 semantic classes are utilized for evaluation. Note that we do not use sequence 08 for model fine-tuning to achieve higher performance on the benchmark. We combine our TransUPR with existing state-of-the-art image-based methods to demonstrate the efficiency and generality of our proposed uncertain point refiner. For the existing state-of-the-art methods, we utilize the available pre-trained models without fine-tuning by validation set and freeze the model weights when we train our refiner. We resubmit the predictions of the baseline methods and methods with our TransUPR to the Semantic KITTI benchmark for comparison. 

We train our TransUPR for 50 epochs regardless of the CNN backbones. We quantitatively evaluate the results of the methods involved via mean intersection-over-union (mIoU), which can be formulated as $mIoU = \frac{1}{C}\sum_{c=1}^{C}\frac{TP_{c}}{FN_{c} + FP_{c} + TP_{c}}$, where $TP_c$ is the number of the true positive cases of the $c^{th}$ class, $FN_c$ is the number of the false negative cases of the $c^{th}$ class, and $FP_c$ is the number of the false positive cases of the $c^{th}$ class. The minor evaluation metric overall accuracy (oACC) is defined as $oACC = \frac{TP}{TP + FP}$.
 
\subsection{Experiment Results}

\begin{table}[t]
    \centering
    \begin{tabular}{c|c|c}
    \hline
         Methods&  oACC & mIoU\\
         \hline\hline
         SalsaNext & 89.3 & 59.0 \\
        SalsaNext + TransUPR & \cellcolor{green}{89.9} & \cellcolor{green}{60.6} \\
        FIDNet & 90.2 & 58.6 \\
        FIDNet + TransUPR & \cellcolor{green}90.5 & \cellcolor{green}59.8 \\
        CENet512 & 88.6 & 61.1 \\
        CENet512 + TransUPR & \cellcolor{green}89.1 & \cellcolor{green}62.9 \\
        CENet1024 & 90.6 & 63.3 \\
        CENet1024 + TransUPR & 90.6 & \cellcolor{green}64.0 \\
        CENet2048 & 90.8 & 63.2 \\
        CENet2048 + TransUPR & \cellcolor{green}\textbf{90.9} & \cellcolor{green}\textbf{64.4} \\
         \hline
    \end{tabular}
    \caption{Quantitative Results Comparison on Semantic KITTI Validation Set (Sequence 08). For the baseline methods without TransUPR, the GPU-enabled KNN refiner is the default post-processing method. (Note: \colorbox{green}{Improved mIoU or oACC} and \textbf{best performance}). }
    \label{tab:2}
\end{table}

As results shown in Table \ref{tab:1}, existing image-based PCSS methods can achieve obvious improvements of mIoU without any architecture modifications via our proposed TransUPR. By combining CENet-512 \cite{cenet} with our TransUPR, we achieve the state-of-the-art performance, i.e., $68.2\%$ mIoU, for image-based methods, which is $+0.6\%$ in mIoU and $+0.4\%$ in oACC over the original state-of-the-art CENet-512\cite{cenet}. For the detailed class-wise IoU, our method improves the IoU in most of the classes except for \textit{motorcyclist} and \textit{traffic-sign}. Table \ref{tab:2} presents the numeric results of selected methods on the Semantic KITTI \cite{semantickitti} validation set. Note that we are following the training schema provided by \cite{cenet} to obtain models with CENet1024 and CENet2048 by fine-tuning the pretrained model of CENet512 step by step. The pretrained model of CENet512 utilized in Table \ref{tab:2} is trained from scratch by ourselves. All the methods with our TransUPR achieve an average $1.3\%$ improvement on mIoU and an average $0.3\%$ improvement on oACC. Although we do not reproduce the results on the original CENet1024 and CENet2048, which CENet2048 should achieve better mIoU than CENet1024, all the CENet models achieve obvious improvement with our proposed TransUPR. Moreover, the CENet-512 with our TransUPR still can achieve about 19 FPS on a single Nvidia RTX 3090, which is much faster than the scanning frequency, i.e., 10Hz, of LiDAR sensors. Fig. \ref{fig:p3} shows the visualization results of CENet512 with our TransUPR on the validation sequence 08.

\subsection{Ablation Study}
 
\begin{table}[h]
    \centering
    \begin{tabular}{c|c|c}
    \hline
         Methods&  oACC & mIoU\\
         \hline\hline
         FIDNet + KNN &  90.2 & 58.6 \\
         FIDNet + TransUPR w/o normal & 90.5 & 59.8 \\ 
         FIDNet + TransUPR w/ normal & 90.3 & 59.8 \\
         \hline
    \end{tabular}
    \caption{Comparison of FIDNet + TransUPR with or without normal vectors on Semantic KITTI sequence 08. }
    \label{tab:3}
\end{table}

\begin{table}[h]
    \centering
    \begin{tabular}{c|c|c}
    \hline
         Methods&  oACC & mIoU\\
         \hline\hline
         SalsaNext & 89.3 & 59.0 \\
        SalsaNext + TransUPR  $c_u=1$  & 89.9 & 60.6 \\
        SalsaNext + TransUPR  $c_u=3$  & 89.9 & 60.5 \\
        SalsaNext + TransUPR  $c_u=4$  & 89.9 & 60.4 \\
         \hline
    \end{tabular}
    \caption{Quantitative Results to explore the influence of $c_u$ for uncertain point selection on Semantic KITTI sequence 08. }
    \label{tab:4}
\end{table}
Other than the geometry features discussed in \ref{sec:2.1}, FIDNet \cite{fidnet} calculates a normal vector for each point and concatenates the five-channel range image and normal vectors as inputs for CNN backbones. Thus, we explore the impact of normal vectors for training our TransUPR, and the quantitative results are shown in Table \ref{tab:3}. The normal vectors for selected uncertain points do not influence the performance of our TransUPR. Both experiments with TransUPR can achieve a 1.3\% improvement in mIoU on sequence 08 compared to the original FIDNet \cite{fidnet}.

Then, we explore the influence of the distance cutoff $c_u$ defined in \ref{sec:2.2} during inference with our TransUPR. Note that we make the $c_u = 1$ for the training of TransUPR to generate more uncertain points. As Table \ref{tab:4} shows, as the $c_u$ increases, the mIoU of the methods with TransUPR decreases, which means only background points have a distance lower than $1$ between its corresponding foreground point, the back-projected label of the background point has high confidence without needing refinement.

\section{Conclusion}

In this paper, we propose a transformer-based plug-and-play uncertain point refiner for LiDAR PCSS, i.e., TransUPR, to improve the semantic segmentation performance of existing image-based LiDAR PCSS approaches. We focus on solving boundary-blurring problems of CNNs and the quantitation loss of spherical projection by applying a learnable and coarse semantic segmentation-guided refiner to replace the common non-learnable KNN refiner. By selecting a limited number of uncertain points for the refiner, we alleviate the harsh memory and computation requirements needed by common transformer networks for LiDAR point cloud understanding. Our TransUPR with CENet outperforms the existing methods and achieves state-of-the-art performance, i.e., 68.2\% mIoU, with an $0.6\%$ mIoU improvement on challenging Semantic KITTI. All the involved methods achieve an average $1.3\%$ mIoU improvement and $0.3\%$ oACC improvement on the Semantic KITTI validation set. Extensive experiments demonstrate that our method is generic and easy to use to combine with any existing image-based LiDAR PCSS approaches. 

\bibliographystyle{IEEEtran}
\bibliography{IEEEbcpat}

\begin{thebibliography}{10}
\providecommand{\url}[1]{#1}
\csname url@rmstyle\endcsname
\providecommand{\newblock}{\relax}
\providecommand{\bibinfo}[2]{#2}
\providecommand\BIBentrySTDinterwordspacing{\spaceskip=0pt\relax}
\providecommand\BIBentryALTinterwordstretchfactor{4}
\providecommand\BIBentryALTinterwordspacing{\spaceskip=\fontdimen2\font plus
\BIBentryALTinterwordstretchfactor\fontdimen3\font minus
  \fontdimen4\font\relax}
\providecommand\BIBforeignlanguage[2]{{%
\expandafter\ifx\csname l@#1\endcsname\relax
\typeout{** WARNING: IEEEtran.bst: No hyphenation pattern has been}%
\typeout{** loaded for the language `#1'. Using the pattern for}%
\typeout{** the default language instead.}%
\else
\language=\csname l@#1\endcsname
\fi
#2}}

\bibitem{squeezeseg}
B.~Wu, A.~Wan, X.~Yue, and K.~Keutzer, ``Squeezeseg: Convolutional neural nets
  with recurrent crf for real-time road-object segmentation from 3d lidar point
  cloud,'' in \emph{2018 IEEE International Conference on Robotics and
  Automation (ICRA)}, 2018, pp. 1887--1893.

\bibitem{rangenet++}
A.~Milioto, I.~Vizzo, J.~Behley, and C.~Stachniss, ``Rangenet ++: Fast and
  accurate lidar semantic segmentation,'' in \emph{2019 IEEE/RSJ International
  Conference on Intelligent Robots and Systems (IROS)}, 2019, pp. 4213--4220.

\bibitem{squeezesegv2}
B.~Wu, X.~Zhou, S.~Zhao, X.~Yue, and K.~Keutzer, ``Squeezesegv2: Improved model
  structure and unsupervised domain adaptation for road-object segmentation
  from a lidar point cloud,'' in \emph{2019 International Conference on
  Robotics and Automation (ICRA)}, 2019, pp. 4376--4382.

\bibitem{squeezesegv3}
C.~Xu, B.~Wu, Z.~Wang, W.~Zhan, P.~Vajda, K.~Keutzer, and M.~Tomizuka,
  ``Squeezesegv3: Spatially-adaptive convolution for efficient point-cloud
  segmentation,'' in \emph{Computer Vision – ECCV 2020}, 2020, p. 1–19.

\bibitem{salsanext}
T.~Cortinhal, G.~Tzelepis, and E.~E.~Aksoy, ``Salsanext: Fast,
  uncertainty-aware semantic segmentation of lidar point clouds,'' in
  \emph{Advances in Visual Computing: 15th International Symposium, ISVC 2020},
  2020, p. 207–222.

\bibitem{pointnet++}
C.~R. Qi, L.~Yi, H.~Su, and L.~J. Guibas, ``Pointnet++: Deep hierarchical
  feature learning on point sets in a metric space,'' in \emph{Advances in
  Neural Information Processing Systems}, vol.~30, 2017.

\bibitem{randla}
Q.~Hu, B.~Yang, L.~Xie, S.~Rosa, Y.~Guo, Z.~Wang, N.~Trigoni, and A.~Markham,
  ``Randla-net: Efficient semantic segmentation of large-scale point clouds,''
  in \emph{2020 IEEE/CVF Conference on Computer Vision and Pattern Recognition
  (CVPR)}, 2020, pp. 11\,105--11\,114.

\bibitem{kpconv}
H.~Thomas, C.~R. Qi, J.~Deschaud, B.~Marcotegui, F.~Goulette, and L.~Guibas,
  ``Kpconv: Flexible and deformable convolution for point clouds,'' in
  \emph{2019 IEEE/CVF International Conference on Computer Vision (ICCV)},
  2019, pp. 6410--6419.

\bibitem{cylinder3d}
X.~Zhu, H.~Zhou, T.~Wang, F.~Hong, Y.~Ma, W.~Li, H.~Li, and D.~Lin,
  ``Cylindrical and asymmetrical 3d convolution networks for lidar
  segmentation,'' in \emph{IEEE Conference on Computer Vision and Pattern
  Recognition}, 2021, pp. 9939--9948.

\bibitem{point2voxel}
Y.~Hou, X.~Zhu, Y.~Ma, C.~C. Loy, and Y.~Li, ``Point-to-voxel knowledge
  distillation for lidar semantic segmentation,'' in \emph{IEEE Conference on
  Computer Vision and Pattern Recognition}, 2022, pp. 8479--8488.

\bibitem{minkowski}
C.~Choy, J.~Gwak, and S.~Savarese, ``4d spatio-temporal convnets: Minkowski
  convolutional neural networks,'' in \emph{2019 IEEE/CVF Conference on
  Computer Vision and Pattern Recognition (CVPR)}, 2019, pp. 3070--3079.

\bibitem{STPLS3D}
M.~Chen, Q.~Hu, H.~Thomas, A.~Feng, Y.~Hou, K.~McCullough, and L.~Soibelman,
  ``Stpls3d: A large-scale synthetic and real aerial photogrammetry 3d point
  cloud dataset,'' in \emph{British Machine Vision Conference}, 2022.

\bibitem{fidnet}
Y.~Zhao, L.~Bai, and X.~Huang, ``Fidnet: Lidar point cloud semantic
  segmentation with fully interpolation decoding,'' in \emph{2021 IEEE/RSJ
  International Conference on Intelligent Robots and Systems (IROS)}, 2021, pp.
  4453--4458.

\bibitem{cenet}
H.~Cheng, X.~Han, and G.~Xiao, ``Cenet: Toward concise and efficient lidar
  semantic segmentation for autonomous driving,'' in \emph{2022 IEEE
  International Conference on Multimedia and Expo (ICME)}, 2022, pp. 01--06.

\bibitem{lite}
R.~Razani, R.~Cheng, E.~Taghavi, and L.~Bingbing, ``Lite-hdseg: Lidar semantic
  segmentation using lite harmonic dense convolutions,'' in \emph{2021 IEEE
  International Conference on Robotics and Automation (ICRA)}, 2021, pp.
  9550--9556.

\bibitem{rpvnet}
J.~Xu, R.~Zhang, J.~Dou, Y.~Zhu, J.~Sun, and S.~Pu, ``Rpvnet: A deep and
  efficient range-point-voxel fusion network for lidar point cloud
  segmentation,'' \emph{2021 IEEE/CVF International Conference on Computer
  Vision (ICCV)}, pp. 16\,004--16\,013, 2021.

\bibitem{fusionnet}
F.~Zhang, J.~Fang, B.~Wah, and P.~Torr, ``Deep fusionnet for point cloud
  semantic segmentation,'' in \emph{Computer Vision -- ECCV 2020}, 2020, pp.
  644--663.

\bibitem{3dmininet}
I.~Alonso, L.~Riazuelo, L.~Montesano, and A.~C. Murillo, ``3d-mininet: Learning
  a 2d representation from point clouds for fast and efficient 3d lidar
  semantic segmentation,'' \emph{IEEE Robotics and Automation Letters}, vol.~5,
  no.~4, pp. 5432--5439, 2020.

\bibitem{af2-s3net}
R.~Cheng, R.~Razani, E.~Taghavi, E.~Li, and B.~Liu, ``(af)2-s3net: Attentive
  feature fusion with adaptive feature selection for sparse semantic
  segmentation network,'' in \emph{Proceedings of the IEEE/CVF Conference on
  Computer Vision and Pattern Recognition (CVPR)}, June 2021, pp.
  12\,547--12\,556.

\bibitem{semantickitti}
J.~Behley, M.~Garbade, A.~Milioto, J.~Quenzel, S.~Behnke, C.~Stachniss, and
  J.~Gall, ``{SemanticKITTI: A Dataset for Semantic Scene Understanding of
  LiDAR Sequences},'' in \emph{Proc. of the IEEE/CVF International Conf.~on
  Computer Vision (ICCV)}, 2019.

\bibitem{calib}
Z.~Zhang, Z.~Yu, S.~You, R.~Rao, S.~Agarwal, and F.~Ren, ``Enhanced
  low-resolution lidar-camera calibration via depth interpolation and
  supervised contrastive learning,'' in \emph{ICASSP 2023 - 2023 IEEE
  International Conference on Acoustics, Speech and Signal Processing
  (ICASSP)}, 2023, pp. 1--5.

\bibitem{squeezenet}
F.~N. Iandola, M.~W. Moskewicz, K.~Ashraf, S.~Han, W.~J. Dally, and K.~Keutzer,
  ``Squeezenet: Alexnet-level accuracy with 50x fewer parameters and
  {\textless}1mb model size,'' \emph{CoRR}, vol. abs/1602.07360, 2016.

\bibitem{resnet}
K.~He, X.~Zhang, S.~Ren, and J.~Sun, ``Deep residual learning for image
  recognition,'' in \emph{2016 IEEE Conference on Computer Vision and Pattern
  Recognition (CVPR)}, 2016, pp. 770--778.

\bibitem{pointrend}
A.~Kirillov, Y.~Wu, K.~He, and R.~Girshick, ``Pointrend: Image segmentation as
  rendering,'' in \emph{2020 CVPR}, 2020, pp. 9796--9805.

\bibitem{transformer}
A.~Vaswani, N.~Shazeer, N.~Parmar, J.~Uszkoreit, L.~Jones, A.~N. Gomez, L.~.
  Kaiser, and I.~Polosukhin, ``Attention is all you need,'' in \emph{Advances
  in Neural Information Processing Systems}, I.~Guyon, U.~V. Luxburg,
  S.~Bengio, H.~Wallach, R.~Fergus, S.~Vishwanathan, and R.~Garnett, Eds.,
  vol.~30, 2017.

\bibitem{pointtransformer}
H.~Zhao, L.~Jiang, J.~Jia, P.~Torr, and V.~Koltun, ``Point transformer,'' in
  \emph{2021 ICCV}, 2021, pp. 16\,239--16\,248.

\bibitem{lovasz}
M.~Berman, A.~R.~Triki, and M.~B. Blaschko, ``The lov{\'a}sz-softmax loss: A
  tractable surrogate for the optimization of the intersection-over-union
  measure in neural networks,'' in \emph{Proceedings of the IEEE Conference on
  Computer Vision and Pattern Recognition}, 2018, pp. 4413--4421.

\end{thebibliography}

\end{document}